\documentclass[dvipsnames,format=sigconf]{acmart}
\AtBeginDocument{%
  }

\copyrightyear{2026}
\acmYear{2026}
\setcopyright{cc}
\setcctype{by}
\acmConference[{\upshape This is the author's version of the work. It is posted here for your personal use. Not for redistribution. The definitive Version of Record was published in} GECCO '26]{Genetic and Evolutionary Computation Conference}{July 13--17, 2026}{San Jos\'e, Costa Rica}
\acmBooktitle{Genetic and Evolutionary Computation Conference (GECCO '26), July 13--17, 2026, San Jos\'e, Costa Rica}
\acmDOI{10.1145/3795095.3805065}
\acmISBN{979-8-4007-2487-9/2026/07}

\usepackage{algorithm}
\usepackage{algpseudocode}
\usepackage{amsmath}
\usepackage{array}
\usepackage{enumitem}
\usepackage{makecell}
\usepackage{multirow}
\usepackage{siunitx}

\newcommand{\argmax}{\operatornamewithlimits{arg\,max}}
\newcommand{\argmin}{\operatornamewithlimits{arg\,min}}
\newcommand{\Cond}{\operatorname{Cond}}
\newcommand{\diag}{\operatorname{diag}}
\newcommand{\mirror}{\operatorname{mirror}}
\newcommand{\modop}{\operatorname{mod}}

\newcommand{\cbox}[2]{\multirow{#1}{*}{\parbox{2.6cm}{\centering #2}}}

\newcommand{\BB}{\boldsymbol{B}}
\newcommand{\eell}{\boldsymbol{\ell}}
\newcommand{\ny}{N_{\omega}}
\newcommand{\mm}{\boldsymbol{m}}
\newcommand{\mt}{\boldsymbol{\tilde{m}}}
\newcommand{\uu}{\boldsymbol{u}}
\newcommand{\xs}{\boldsymbol{x^*}}
\newcommand{\qq}{\boldsymbol{q}}
\newcommand{\xx}{\boldsymbol{x}}

\newcommand{\yy}{\boldsymbol{y}}
\newcommand{\zz}{\boldsymbol{z}}
\newcommand{\yh}{\boldsymbol{\hat{y}}}
\newcommand{\yhp}{\boldsymbol{\hat{y}}^{\boldsymbol{\prime}}}
\newcommand{\ys}{\boldsymbol{y^*}}
\newcommand{\ysx}{\boldsymbol{y^*_x}}
\newcommand{\ysxs}{\boldsymbol{y^*_{x^*}}}

\newcommand{\DD}{\boldsymbol{D}}
\newcommand{\SSigma}{\boldsymbol{\Sigma}}
\newcommand{\SSigmat}{\boldsymbol{\tilde{\Sigma}}}
\newcommand{\T}{^{\top}}
\newcommand{\Tmin}{T_{\mathrm{min}}}
\newcommand{\ind}[1]{\mathbb{I}\{#1\}} %

\newcommand{\ttheta}{\boldsymbol{\theta}}
\newcommand{\tthetat}{\boldsymbol{\tilde{\theta}}}
\newcommand{\oomega}{\boldsymbol{\omega}}
\newcommand{\oomegat}{\boldsymbol{\tilde{\omega}}}

\newcommand{\tthr}{\tau_{\mathrm{threshold}}}
\newcommand{\pthr}{p_{\mathrm{threshold}}}
\newcommand{\pp}{\bar{p}_{+}}
\newcommand{\pn}{\bar{p}_{-}}

\newcommand{\Vxmin}{V^{x}_{\mathrm{min}}}
\newcommand{\Vymin}{V^{y}_{\mathrm{min}}}
\newcommand{\Cxmax}{\mathrm{Cond}_{\mathrm{max}}^{x}}
\newcommand{\Cymax}{\mathrm{Cond}_{\mathrm{max}}^{y}}
\newcommand{\cmax}{c_{\mathrm{max}}}

\newcommand{\ph}{$\phantom{<}$} \begin{document}

\title[Accelerating Black-Box Bilevel Optimization with Rank-Based Upper-Level Value
Function Approximation]{  %
    Accelerating Black-Box Bilevel Optimization with\\
    Rank-Based Upper-Level Value Function Approximation}

\author{Marc Ong}
\email{marc@bbo.cs.tsukuba.ac.jp}
\orcid{0009-0009-9667-3026}
\affiliation{%
  \institution{University of Tsukuba}
  \city{Tsukuba}
  \state{Ibaraki}
  \country{JAPAN}
}

\author{Youhei Akimoto}
\email{akimoto@cs.tsukuba.ac.jp}
\orcid{0000-0003-2760-8123}
\affiliation{%
  \institution{University of Tsukuba \& RIKEN AIP}
  \city{Tsukuba}
  \state{Ibaraki}
  \country{JAPAN}
}

\begin{abstract}
Bilevel optimization is a field of significant theoretical and practical interest, yet solving such optimization problems remains challenging. Evolutionary methods have been employed to address these problems in the black-box setting; however, they incur high computational cost due to the nested nature of bilevel optimization. Although previous methods have attempted to reduce this cost through various heuristic techniques, such approaches limit versatility on challenging optimization landscapes, such as those with multimodality and significant interaction between upper- and lower-level decision variables. In this study, we propose an efficient framework that exploits the invariance of rank-based evolutionary algorithms to monotonic transformations, thereby reducing the computational burden of the lower-level optimization loop. Specifically, our method directly approximates the rankings of the upper-level value function, bypassing the need to run the lower-level optimizer until convergence for each upper-level iteration. We apply this framework to the setting where both levels are continuous, adopting CMA-ES as the optimizer. We demonstrate that our method achieves competitive performance on standard bilevel optimization benchmarks and can solve problems that are intractable with previously proposed methods, particularly those with multimodality and strong inter-variable interactions.
\end{abstract}



\maketitle

\section{Introduction}
Bilevel optimization problems (BLOPs) are a class of problems in which one optimization problem is nested within another, and can be expressed in the following general form:
\begin{equation}
    \begin{aligned}
    \min_{\xx, \ysx} \quad & F(\xx, \ysx) \\ 
    \mathrm{s.t.} \quad & \ysx \in \argmin_{\yy} f(\xx, \yy).
    \end{aligned}
\end{equation}
Here, $\xx \in \mathbb{X}$ and $\yy \in \mathbb{Y}$ denote the
upper- and lower-level decision variables; $F:\mathbb{X}
\times \mathbb{Y} \to \mathbb{R}$ and $f:\mathbb{X} \times
\mathbb{Y} \to \mathbb{R}$ denote the upper- and lower-level objectives.\footnote{In general, the upper-level minimization may require selecting from multiple possible values of $\ysx$ if multiple lower-level minima exist, in addition to optimizing $\xx$.}

This hierarchical structure is intrinsic to many real-world problems, leading to
applications across diverse fields such as logistics~\cite{sun2008}, supply
chain optimization~\cite{ma2014}, energy systems~\cite{xiao2018}, mobile
networks~\cite{chen2024}, and various areas of machine learning, including
hyperparameter optimization~\cite{franceschi2018,dagreou2022}, neural
architecture search~\cite{cui2019,he2020}, and behavioral
alignment~\cite{gupta2023,thoma2024}. Despite their ubiquity, however, BLOPs are
challenging due to the high computational cost incurred by the nested structure;
even simple linear BLOPs are NP-hard~\cite{jeroslow1985}. The
difficulty is further exacerbated in the black-box setting, where structural
information that might otherwise simplify the problem is unavailable. In
practice, the cost of repeatedly solving the lower-level problem to convergence
for each upper-level iteration is often prohibitive, particularly when the two
levels interact strongly~\cite{he2018}. This makes the lower level a significant bottleneck in
terms of both computational cost and scalability.

Methods for solving black-box BLOPs fall into three main categories:
single-level reduction, nested, and approximation-based. Single-level reduction
methods use lower-level optimality conditions to reformulate the problem into a
single level, and utilize methods such as genetic
algorithms~\cite{hejazi2002} or particle swarm optimization~\cite{jiang2013} to
optimize the resulting single-level problem. They are useful in certain
situations where these conditions are known and possess nice mathematical
properties, but are largely unsuitable for black-box settings where  such conditions
are unavailable. Nested methods, such as BIDE~\cite{angelo2013} and
NBLEA~\cite{sinha2014}, are more general, employing an upper-level optimizer that
repeatedly calls a lower-level solver following the nested structure of the
BLOP. This generality makes them suitable for black-box applications; however,
this approach incurs significant computational cost due to frequent calls to the lower-level solver. Lastly, approximation-based methods
solve a simplified problem, either by directly approximating the lower-level
objective, as in BLEAQ~\cite{sinha2017}, or by constructing surrogate models, as in
SABLA~\cite{islam2017} and BLEAQ2~\cite{sinha2020}, though challenges with surrogate
construction and scalability persist.

This study focuses on the continuous domain for both the upper and lower levels, where $\mathbb{X}=\mathbb{R}^{d_x}$ and $\mathbb{Y}=\mathbb{R}^{d_y}$. In such cases, the CMA-ES algorithm~\cite{hansen2001,hansen2003,akimoto2020} is considered a state-of-the art optimizer due to several of its attractive
properties, including various invariance guarantees and automatic adaptation of
mutation strength. It is suitable for solving problems with challenging
properties, such as ill-conditioning, nonconvexity, nonseparability, and
nonsmoothness~\cite{hansen2010}, making it desirable as a building block within
bilevel optimization algorithms. BL-CMA-ES~\cite{he2018} was the first to apply
CMA-ES to BLOPs in the black-box setting, employing it for the optimization at
both levels within the nested approach, and showed improved performance and
efficiency compared to previous methods on standard benchmarks. Subsequent
research has built on the BL-CMA-ES framework to overcome limitations of the
original implementation~\cite{chen2021,huang2023,xu2025}, achieving various speed-ups.

\subsection{Related Work}
To contextualize our contributions, we provide a brief chronology of notable
applications of evolution strategies to bilevel optimization leading up to our
proposed contribution. 

\subsubsection{BL-CMA-ES}
To address shortcomings of previous methods, He et al.~\cite{he2018}
introduced BL-CMA-ES, the first CMA-ES-based algorithm for BLOPs, forming the
basis for future approaches. At the upper level, the
algorithm uses a single CMA-ES solver that learns a joint distribution over
upper- and lower-level decision variables. At each step, it samples pairs $(\xx,
\yy)$ and, for each pair, a temporary lower-level CMA-ES solver is initialized
using a marginal distribution of $\yy$ computed from the upper-level 
joint distribution, and produces an estimate $\yh$ of the lower-level optimum with $\xx$
fixed. The resulting pairs $(\xx, \yh)$ are then ranked according to the
upper-level objective to update the upper-level solver, and this procedure is
repeated until a convergence criterion is met.

On preliminary benchmarks, BL-CMA-ES was shown to be competitive against BLEAQ2,
SABLA, NBLEA and BIDE. Despite these advances, several weaknesses limit the
approach's efficiency and applicability. These can be broadly described as
follows:

\begin{enumerate}
  \item\label{lim:mismatch} \textit{Sampling and update distributions mismatch:}
  The algorithm updates the upper-level distribution using refined pairs $(\xx,
  \yh)$, which differ from the originally sampled pairs $(\xx, \yy)$. This
  discrepancy violates a design assumption of the CMA-ES algorithm~\cite{hansen2011} and may hinder
  convergence.
  
  \item\label{lim:init} \textit{Initialization can be ineffective:} For problems
  with strong interaction between $\xx$ and $\yy$, initialization of the
  lower-level distribution based on marginalization can be counterproductive or
  misleading to the lower-level solver, increasing the computational cost. 
  
  \item\label{lim:discard} \textit{Information is inefficiently discarded:}
  Knowledge gained during the lower-level optimization is thrown away after each
  upper-level update.
  Instead, the next generation of lower-level solvers must use the potentially
  unhelpful marginalized upper-level distribution as its starting point.
  
  \item\label{lim:collab} \textit{Solvers do not collaborate:} Similarly, there
  is no mechanism allowing information to propagate across lower-level solvers
  corresponding to different upper-level candidates. This prevents potential
  speed-ups when these candidates are near each other.
\end{enumerate}

Limitation~\ref{lim:mismatch} causes issues with convergence in difficult
upper-level landscapes. The authors of BL-CMA-ES partially mitigate this issue
by using an elite preservation mechanism to save older solutions in case the
performance of new candidates degrades, but a distributionally correct approach
would likely offer a more robust remedy to this problem. 

The remaining three limitations are associated with slow lower-level
optimization, exacerbating the main cause of the high computational cost of
nested approaches. Addressing all of these limitations could lead to improved
lower-level convergence speed; additionally, resolving Limitation~\ref{lim:init}
provides the benefit of robustness in situations with strong inter-variable
interactions.
 
\subsubsection{TL-CMA-ES and BOC}
Subsequent research has aimed to improve the flexibility and performance of the
BL-CMA-ES framework, primarily focusing on improving lower level search. For
example, TL-CMA-ES, proposed by Chen and Liu~\cite{chen2021}, incorporates a
transfer learning mechanism that allows lower-level solvers with nearest
neighbor upper-level solution candidates in Euclidean distance to share
parameters. The BOC method~\cite{huang2023} builds on this with a collaboration
mechanism that enables the direct sharing of candidate solutions within the lower-level optimization loop, outperforming both BL-CMA-ES
and TL-CMA-ES on preliminary benchmarks.

While recent advances~\cite{chen2021,huang2023,xu2025} have primarily focused on
implementing collaboration mechanisms to address Limitation~\ref{lim:collab},
the other limitations of BL-CMA-ES-based approaches remain unaddressed. A
collaboration mechanism such as that in BOC can be effective when lower-level
search distributions are similar across upper-level candidates; however, it has
notable drawbacks. First, because BOC shares a single lower-level population
across all solvers, it lacks the search capacity of BL-CMA-ES. Moreover, BOC
provides no mechanism for judging the appropriateness of shared information. We
expect these drawbacks to cause difficulties on problems with multimodality or
strong interaction between levels.
Consequently, we expect the attempt to resolve
Limitation~\ref{lim:collab} via collaboration to create a tradeoff between higher efficiency and
lower versatility compared to the original BL-CMA-ES.

\subsubsection{Evolutionary approaches for min-max optimization}
On the other hand, recent literature has investigated solving min-max
problems in the black-box setting, particularly in the context of robust
simulation-based optimization. Min-max problems are those of the form
\begin{equation}
    \min_{\xx} \max_{\yy} f(\xx,\yy),
\end{equation}
and can be viewed as a special case of bilevel optimization. Indeed, as Antoniou
and Papa~\cite{antoniou2020} demonstrate, BL-CMA-ES can be directly
applied to min-max problems. However, they also remark that further research
into the applicability of bilevel methods to min-max problems, and the crossover
between these two areas more broadly, is still needed.

Existing evolutionary methods for min-max optimization include Adversarial
CMA-ES~\cite{akimoto2022} and WRA-CMA-ES~\cite{miyagi2023}. Adversarial CMA-ES adopts a
simultaneous ascent-descent approach, which is unsuitable for BLOPs.
In contrast, WRA-CMA-ES employs a nested structure: an outer loop minimizes
$f(\xx, \yh(\xx))$ over $\xx$, while an inner loop performs the maximization to
approximate $\yh(\xx) \approx \argmax_{\yy} f(\xx, \yy)$ for each $\xx$.
WRA-CMA-ES introduces two innovations, warm starting and early stopping, that
effectively address the other limitations of BL-CMA-ES and the high
computational cost of the nested loop. We summarize these contributions as
follows:

\begin{itemize}
  \item \textbf{Warm starting.} The algorithm maintains a cache of $\ny$
  different configurations for the lower-level solver. Prior to initializing the
  lower-level CMA-ES instances, the best configuration $k$ for each upper-level
  candidate, as evaluated by $f(\xx,\yy_k)$, is selected, and this cache is
  updated after the lower-level optimization is complete.
  
  \item \textbf{Early stopping.} As CMA-ES is invariant to monotonic
  transformations and depends only on rankings, the estimate of $\yh$ can be
  considered sufficiently approximated if its rankings, when evaluated by $f$,
  do not change. This helps reduce the high computational cost of the
  inner optimization over $\yh$.
\end{itemize}
We believe that these innovations also provide hints for improving 
general black-box bilevel optimization algorithms, including BL-CMA-ES and its derivatives.

\subsection{Our Contribution}
The nested architecture of WRA-CMA-ES is well-suited for general bilevel
optimization.
Specifically, its lower-level solvers can be modified to approximate the lower-level optimal response $\ysx$ for each upper-level candidate solution $\xx$, while 
its upper-level solver approximates the upper-level value
function $\Phi(\xx) \coloneq F(\xx, \ysx)$, provided that the optimization problem
does not require selecting from a multivalued set of lower-level solutions
$\ysx$. In this generalization, the upper level both samples from and
updates parameters according to the same distribution over $\Phi$, eliminating the mismatch described in Limitation~\ref{lim:mismatch}. Moreover, the
warm-starting and early-stopping strategies may provide benefit to the other
aforementioned limitations of previous approaches, providing both selective
information sharing and acceleration of the lower-level search.

We propose Upper-Level Ranking Approximation CMA-ES (URA-CMA-ES), a generalization of the
WRA-CMA-ES method to bilevel optimization that incorporates warm-starting and
early-stopping strategies. These features aim to overcome the limitations of
previous methods at both the upper and lower levels. We hypothesize that the
proposed method's selective information sharing provides an
acceleration over BL-CMA-ES comparable to that of BOC, while avoiding BOC's
weaknesses to multimodality. Moreover, we expect
that overcoming the other limitations will improve versatility over both BOC
and BL-CMA-ES on functions with strong inter-variable interaction.

To verify our hypotheses, we benchmark URA-CMA-ES against BOC and BL-CMA-ES on
two standard test suites: the SMD suite~\cite{sinha2014}, which is commonly used
in the bilevel optimization community, and the WRA suite~\cite{miyagi2023} of
min-max optimization problems. These benchmarks were selected because they
contain problems with the strong interaction and
multimodality required to test our claims, in addition to other difficult
characteristics such as ill-conditioning, nonconvexity and nonsmoothness.

\section{Proposed Method}
Similar to WRA-CMA-ES, URA-CMA-ES uses a nested optimization approach. An outer CMA-ES solver minimizes an estimate of the upper-level value function $\tilde{\Phi}(\xx)\coloneq F(\xx,\yh(\xx))$, where the estimated lower-level optimal response $\yh(\xx) \approx \argmin_{\yy} f(\xx, \yy)$ for a given upper-level candidate solution $\xx$ is obtained via an inner solver.
This contrasts with the original WRA-CMA-ES, which optimizes the same objective at both levels, minimizing over $\xx$ using an outer solver and maximizing over $\yy$ for a given $\xx$ using an inner solver. In this work, we treat the lower level as a black box, making CMA-ES a natural choice for these inner solvers as well, though any suitable solver can be easily substituted depending on the information available about the lower-level problem.

Moreover, we design URA-CMA-ES to inherit the warm-starting and early-stopping
mechanisms of WRA-CMA-ES. We next discuss the details and expected effects of
these mechanisms by introducing the specifics of the upper-level optimization
and upper-level ranking approximation procedures.

\paragraph{Upper-level optimization}
The upper-level loop begins with the CMA-ES solver sampling $\lambda_x$
upper-level candidate solutions $\{\xx_i\}_{i=1}^{\lambda_x}$. These
solutions are then evaluated using the upper-level ranking approximation~(URA)
procedure, as outlined in Algorithm~\ref{alg:WRA}, to estimate their upper-level
objective values. Subsequently, the candidates are ranked based on these
approximate values, and this ranking is used to update the solver's distribution
parameters. Specifically, the update rule adopted by the dd-CMA-ES variant is
used~\cite{akimoto2020}, with diagonal decoding acceleration disabled for fairness of
comparison to previous methods. As this upper-level CMA-ES solver both samples
and updates according to the same approximated upper-level value function, this addresses
the distribution mismatch issue described in Limitation~\ref{lim:mismatch}.

This iterative process continues until termination criteria based on the
covariance matrix $\SSigma_x$ are met. Termination occurs when the maximum
coordinate-wise standard deviation, i.e. the square root of the largest diagonal
element $\max_{\ell\in\{1,\ldots,d_x\}}\sqrt{[\SSigma_x]_{\ell,\ell}}$, falls
below a threshold hyperparameter $\Vxmin$, or the condition number, i.e. the
ratio of the largest and smallest eigenvalues of $\SSigma_x$, exceeds a
hyperparameter $\Cxmax$.

\begin{algorithm}[t]
    \caption{Upper-level ranking approximation}\label{alg:WRA}
    \begin{algorithmic}[1]
    \Require $\xx_1, \dots, \xx_{\lambda_x}$ 
    \Require $\{(\yy_k, \oomega_k, p_k)\}_{k=1}^{\ny}$
    \Require $\tthr$, $\pthr$, $\pp$, $\pn$
    
    \State\label{line:WRA-initstart}\texttt{// 
    Warm starting
    }
    \For{$i \in \{ 1, \dots, \lambda_x\}$}
    \State evaluate $f(\xx_{i}, \yy_k)$ for all $k \in \{1,\dots,\ny\}$
    \State $k^\mathrm{min}_{i} \gets \argmin_{k \in \{ 1, \dots,\ny\}} f(\xx_i, \yy_k)$
    \State $\yh_i\gets\yy_{k^\mathrm{min}_{i}}$, $\oomegat_i\gets\oomega_{k^\mathrm{min}_{i}}$
    \State $\tilde f^0_i\gets f(\xx_i, \yy_{k^\mathrm{min}_{i}})$, $\tilde\Phi^0_i\gets F(\xx_i, \yy_{k^\mathrm{min}_{i}})$ 
    \EndFor\label{line:WRA-initend}
    
    \State\label{line:WRA-worststart}\texttt{// Early stopping}
    \State initialize $\tthetat_1, \dots, \tthetat_{\lambda_x}$
    \For{$\mathrm{rd} \in \{ 1,2,\dots\}$}
    \For{$i \in \{1, \dots, \lambda_x\}$}
    \State $\tilde f_{i}^{\mathrm{rd}}, \yh_i, \oomegat_i, \tthetat_i \gets \mathcal{M}(\tilde f_{i}^{\mathrm{rd}-1}, \yh_i, \oomegat_i, \tthetat_i)$ 
    \EndFor
    \State $\tilde\Phi_i^\mathrm{rd} \gets \{F(\xx_i,\yh_i)\}_{i=1}^{\lambda_x}$%
    \State $\tau \gets \mathrm{Kendall}(\{\tilde\Phi^{\mathrm{rd}-1}_{i}\}_{i=1}^{\lambda_x}, \{\tilde\Phi^{\mathrm{rd}}_{i}\}_{i=1}^{\lambda_x})$
    \State \textbf{break if} $\tau > \tau_{\mathrm{threshold}}$
    \EndFor\label{line:WRA-worstend}
    
    \State\texttt{// Post-processing}\label{line:WRA-poststart}
    \State $S^\mathrm{min} \gets \{{k^\mathrm{min}_{i}}\}_{i=1}^{\lambda_x}$
    \For{$\tilde{k} \in S^\mathrm{min}$}
    \State $\ell \gets \argmin_{i=1,\dots,\lambda_x} \{\tilde\Phi_i \mid k^\mathrm{min}_i = \tilde{k}\}$ 
    \State $\yy_{\tilde{k}} \gets \yh_\ell$, $\oomega_{\tilde{k}} \gets \oomegat_\ell$  
    \State $p_{\tilde{k}} \gets \min(p_{\tilde{k}} + \bar{p}_{+}, 1)$
    \EndFor
    \State $p_k \gets p_k - \bar{p}_{-} \cdot \ind{k \notin S^\mathrm{min}}$ for all $k\in\{1,\ldots,\ny\}$
    \For{$k \in \{1,\dots,\ny\}$}
    \State refresh $(\yy_k, \oomega_k, p_k)$ \textbf{if} $p_k < p_{\mathrm{threshold}}$
    \EndFor
    \label{line:WRA-postend}
    \State \Return $\{\tilde\Phi_i^\mathrm{rd}\}_{i=1}^{\lambda_x}$ and $\{(\yy_k, \oomega_k, p_k)\}_{k=1}^{\ny}$ for the next call
  \end{algorithmic}
\end{algorithm}

\begin{algorithm}[t]
    \caption{CMA-ES as $\mathcal{M}$}\label{alg:CMAESinWRA}
    \begin{algorithmic}[1]%
    \Require $\xx, \yh, f_y, \oomegat = (\mt, \SSigmat), \tthetat = (h, t', \dots)$ 
    \Require $\Vymin >0$, $\cmax \geq 1$, $\lambda_y = \lfloor 4 + 3\log(d_y) \rfloor$
    
    \State $\SSigmat_{\mathrm{init}} \gets \SSigmat$, $c \gets 0$, $h \gets \textsc{False}$
    \While{$c < \cmax$ \textbf{and} $h = \textsc{False}$}%
    
    \State Sample $ \{\yhp\}_{k=1}^{\lambda_y} \sim \mathcal{N}(\mt, \SSigmat)$
    \State Evaluate $f_k \gets f(\xx, \yhp_{k})$ for all $k\in\{1,\ldots,\lambda_y\}$
    
    \State Select the min index $\tilde{k}^{\mathrm{min}}\gets\argmin_{k\in\{1,\dots,\lambda_y\}} f_k$
    \If {$f(\xx, \yhp_{\tilde{k}^{\mathrm{min}}}) \le f_{y}$}
    \State $f_{y} \gets \min_{k\in\{1,\dots,\lambda_y\}} f_k$, $\yh \gets \yhp_{\tilde{k}^{\mathrm{min}}}$, $c \gets c + 1$
    \EndIf
    
    \State Perform CMA-ES update using $\{\yhp_{k}, f_k\}_{k=1}^{\lambda_y}$
    \If {$\max_\ell\sqrt{[\SSigmat]_{\ell,\ell}} < \Vymin$ \text{and} $t' \geq \Tmin$} 
     \State\label{l:sig_correct_1} $\DD \gets \diag\bigg(\max\bigg(1, \frac{V_{\min}^y}{\sqrt{[\SSigmat]_{1,1}}}\bigg), \dots, \max\bigg(1, \frac{V_{\min}^y}{\sqrt{[\SSigmat]_{d_y,d_y}}}\bigg)\bigg)$

     \State\label{l:sig_correct_2} $\SSigmat \gets \DD \SSigmat \DD$ and $h \gets \textsc{True}$
    \EndIf
    \State Set $\SSigmat \gets \SSigmat_{\mathrm{init}}$ and $h \gets \textsc{True}$ \textbf{if} $\Cond(\SSigmat) > \Cymax$
    \State $t' \gets t'+1$
    \EndWhile
    \State \Return $\yh$, $f_y$, $\oomegat = (\mt, \SSigmat)$, $\tthetat = (h, t', \dots)$
  \end{algorithmic}
\end{algorithm}

\paragraph{Upper-level ranking approximation}
As outlined in Algorithm~\ref{alg:WRA}, the algorithm takes as input $\lambda_x$
candidates from the upper-level solver, a cache of $\ny$ configurations from the
previous generation's lower-level solvers, and various hyperparameters. It
returns approximated rankings of the upper-level value function $\tilde\Phi$ and
an updated cache. We use $\oomega$ to denote configurations inherited by the
lower-level solvers (i.e., mean vector and covariance matrix) and $\ttheta$ for
non-inherited settings (e.g., evolution path, termination flag, and iteration
count). The procedure consists of three phases:

  \paragraph{\textup{\textbf{(1) Warm starting.}}} For each of the $\lambda_x$ candidate
  solutions, the best-performing configuration from the cache, as measured by
  the lower-level objective function, is selected to initialize the lower-level
  solver. This allows information to be passed from the previous generation,
  solving the ineffective initialization problem (Limitation~\ref{lim:init}) and
  lack of information sharing mechanism across generations
  (Limitation~\ref{lim:discard}). It also facilitates selective information
  sharing within the same generation (Limitation~\ref{lim:collab}), as lower-level solvers corresponding to
  different candidate $\xx$ values would be initialized with the same
  distribution only if expected to be beneficial. 
  
  \paragraph{\textup{\textbf{(2) Lower-level optimization with early stopping.}}} Each candidate's
  lower-level solver $\mathcal{M}$ is initialized with its warm-started
  configuration and executes the optimization. As described in Algorithm 2, a
  single round runs the lower-level CMA-ES until it either achieves a target
  number of improvements ($\cmax$) or meets specific termination criteria. These
  criteria, which are based on maximum coordinate-wise standard deviation
  falling below $\Vymin$ and the covariance matrix condition number exceeding
  $\Cymax$, are similar to those of the upper-level solver. However, unlike
  in the upper level, the standard deviation is allowed to fall below $\Vymin$
  until $\Tmin$ iterations are counted.
  
  After each round, the estimate of the
  upper-level $\tilde\Phi$ is computed from the approximate optimal
  response $\yh$, and Kendall's rank correlation coefficient measures the change
  in solution rankings when compared with the previous round's estimated upper-level value
  function. If the correlation exceeds a threshold $\tthr$, the optimization is
  considered converged, and the loop terminates early.

  \paragraph{\textup{\textbf{(3) Post-processing.}}} If a solver selected some configuration
  $k\in\{1,\ldots,\ny\}$ during the lower-level optimization, the
  post-processing phase overwrites the $k$-th cache entry with that solver's
  newly learned distribution parameters. If multiple solvers select the same
  configuration, the tie is broken using the estimated 
  $\tilde\Phi(\xx)$ of their corresponding upper-level candidates. Finally,
  to encourage exploration, a refresh mechanism is then applied to the cache.
  Each configuration has a corresponding score $p_k \in[0,1]$, which is incremented
  by $\pp$ when the configuration is selected and decremented by $\pn$ when it is
  not. If $p_k$ drops below a threshold $\pthr$, the configuration is
  re-initialized to a random value. This mechanism also serves as a filter to
  prevent irrelevant information from being transmitted to the next generation
  of lower-level solvers.

\paragraph{Hyperparameters}
We mostly reference the default hyperparameters from the original WRA-CMA-ES
implementation. The settings are organized by algorithm component below.
\begin{itemize}
  \item \textbf{Upper-level CMA-ES:} %
  $\lambda_x=\lfloor4+3\log(d_x)\rfloor$ is used as the upper-level population
  size. The covariance matrix-based termination hyperparameters are
  $\Vxmin=\num{1e-12}$ and $\Cxmax=\num{1e7}$.
  
  \item \textbf{Upper-level ranking approximation (Algorithm~\ref{alg:WRA}):}
  We use $\tthr=0.7$,
  $\pthr=0.1$, $\pp=0.4$, and $\pn=0.05$. The number of configurations $\ny$ is
  set to $3\times\lambda_x$. Scores for all configurations are initialized to $p_k=1$ ($k\in\{1,\ldots,\ny\}$).
  
  \item \textbf{Lower-level optimization (Algorithm~\ref{alg:CMAESinWRA}):}
  For the lower-level CMA-ES hyperparameters, we use $\lambda_y = \lfloor 4 + 3\log(d_y) \rfloor$, $\cmax=1$, $\Tmin=10$, $\Vymin=\num{1e-4}$, and $\Cymax=\num{1e7}$.
\end{itemize}

\section{Experiments}

To evaluate our hypotheses regarding the proposed method's efficiency and
versatility, we benchmark URA-CMA-ES on two main test
suites: SMD and WRA.\footnote{The source code for the implementation and benchmarking of URA-CMA-ES is available at \texttt{https://github.com/akimotolab/ura}.
} For fairness of comparison, we use the recommended
hyperparameters for BL-CMA-ES and BOC proposed by the authors, as well as the
same shared termination conditions and other experimental settings where possible. Following previous literature, we report
medians and interquartile ranges (IQRs) for upper-level accuracy
$|F-F^*|\coloneq|F(\xx,\yy)-F(\xs,\ysxs)|$, lower-level accuracy
$|f-f^*|\coloneq|f(\xx,\yy)-f(\xs,\ysxs)|$, and upper- and lower-level function
evaluations (FEs) across 20 runs for each method. We also perform an ablation study on URA-CMA-ES to clarify the contributions of the early stopping and warm starting mechanisms. %

\subsection{Experimental Settings}\label{ss:exp_settings}
\subsubsection{Initial Distribution Parameters}\label{ss:initialdist}
All benchmark problems in this study specify box constraints, which we denote as $\xx \in [\eell_x,\uu_x]$ and $\yy \in [\eell_y, \uu_y]$. For the proposed method, we initialize the upper-level mean vector as $\mm_x = \eell_x + \mathcal{U}(0,1) \times (\uu_x - \eell_x)$ and the covariance matrix as $\SSigma_x = [\diag((\uu_x-\eell_x)/4)]^2$,
where $\mathcal U(0,1)$ denotes a scalar chosen randomly from a standard uniform distribution. The lower-level parameters are similarly initialized as $\mm_y = \eell_y + \mathcal{U}(0,1) \times (\uu_y - \eell_y)$ and $\SSigma_y = [\diag((\uu_y-\eell_y)/4)]^2$.

As the upper-level optimizers of BL-CMA-ES and BOC learn a joint distribution over $\zz \coloneq [\xx;\yy]$ at the upper level, we initialize their upper-level distribution parameters as $\mm_z = \eell_z + \mathcal{U}(0,1) \times (\uu_z - \eell_z)$ and $\SSigma_z = [\diag((\uu_z-\eell_z)/4)]^2$, where $\eell_z \coloneq [\eell_x; \eell_y]$ and $\uu_z \coloneq [\uu_x; \uu_y]$ are the concatenated box constraint boundaries. Note that with this choice of upper-level initialization, the marginalization mechanism used in BL-CMA-ES and BOC automatically initializes the lower-level distribution parameters identically to the values we set for the proposed method.

\subsubsection{Constraint Handling} At both levels, we adopt a mirroring technique~\cite{yamaguchi2018} to
handle the box constraints.
Specifically, any infeasible solution is mirrored back into the feasible region:
\begin{equation}
    [\mirror(\qq)]_i = u_i - \left| \modop(q_i - \ell_i, 2(u_i - \ell_i)) - (u_i - \ell_i) \right|,
\end{equation}
where $\qq\in[\eell,\uu]$ denotes the decision variable (either $\xx$ or $\yy$). The objective function value of the mirrored solution, $f(\mirror(\qq))$, is then used as the evaluation of the original infeasible solution $\qq$. Note that the CMA-ES update is performed using the original infeasible solutions, and the distribution is allowed to exit the feasible region.

\subsubsection{Budget-Based Termination}
We employ the following termination criteria. The upper level terminates when
its objective value is within $\num{1e-6}$ of the true optimum, or when no
improvement exceeding $\num{1e-6}$ is observed for $60$ consecutive iterations.
To allow convergence, no maximum iteration limit is imposed on the upper level.
The lower level terminates when improvement falls below
$\num{1e-6}$ for $20$ consecutive iterations, with a maximum of $50$ iterations. A restart strategy (Section~\ref{ss:restart}) resets all
parameters to encourage exploration until a total budget of $\num{1e7}$~FEs is
exhausted. Distribution parameters for both initialization and restarts are
randomly generated according to Section~\ref{ss:initialdist}.

\subsubsection{Restart Strategy}\label{ss:restart}
To facilitate exploration, we implement a basic restart strategy. When
an upper-level termination condition is triggered either by the algorithm (i.e.,
based on the properties of the covariance matrix) or the experimental settings, we re-initialize all
parameters of the algorithm according to Section~\ref{ss:initialdist}. This procedure is repeated until
the total FE budget of $\num{1e7}$ is exhausted. Note that
restarting is aborted if the algorithm reaches the required upper-level accuracy
of $\num{1e-6}$ once. The best solution across restarts is recorded and used for the final reported accuracies.

\subsection{Benchmark Problems}\label{ss:benchmarkproblems}
We provide a brief overview of the benchmark test suites used in this study. We
set the dimensionality to $(d_x+d_y)=(20+20)$ for all problems to ensure that
their difficulties are pronounced.
\paragraph{SMD} The unconstrained SMD test suite consists of
eight BLOPs that present various challenges, including ill
conditioning, non-convexity, and multimodality. No constraints are
employed, with the exception of box constraints. These problems are scalable to
arbitrary dimensionality, with the restriction that $d_x
\le d_y$. Note that there is also a constrained SMD test suite; however, in high
dimensions, all of these problems contain exceedingly complex constraints,
rendering them unsolvable using black-box methods~\cite{huang2023}.
\paragraph{WRA}
The WRA test suite consists of 11 min-max optimization problems. These problems
exhibit various challenges, such as strong variable interaction, non-smoothness,
and non-convexity, and are scalable to arbitrary dimensionality in both $\xx$
and $\yy$. The identity matrix is chosen as the interaction matrix
$\BB$, and the default box constraints of $\xx\in[-3,3]^{d_x}$ and
$\yy\in[-3,3]^{d_y}$ are used.

\subsection{Results on SMD}
\begin{table*}
\caption{
\boldmath
Median values and IQRs (in parentheses) of accuracy and function evaluations (FEs)
for the $(20+20)$-dimensional unconstrained SMD suite over 20 runs.
Upper-level accuracies below $1\times10^{-6}$ tolerance are reported as $1\times10^{-6}$ for
ease of comparison. $<$ and $>$ denote FEs significantly less or
greater than URA-CMA-ES, respectively, based on a Wilcoxon signed rank test
with $p<0.05$ and Bonferroni correction; the
absence of a symbol indicates comparison is not possible due to differing success outcomes across methods.
Success denotes convergence in ${>}75\%$ of runs (main value) and ${>}50\%$
(parenthesized value).
}
\label{table:smd}
\centering
{
\begin{tabular}{cccccc}
\bottomrule
Test problem & Comment & & URA-CMA-ES (ours) & BOC \ph & BL-CMA-ES \ph \\ \hline
\multirowcell{4}{SMD1} & \cbox{4}{Cooperative, convex at both levels}
& $|F-F^*|$   & \textbf{1.00E-6 (0.00E+0)} & \textbf{1.00E-6 (0.00E+0)} \ph & \textbf{1.00E-6 (0.00E+0)} \ph \\
& & Upper FEs &          6.31E+3 (2.16E+2) &          7.76E+3 (5.09E+2) $>$ &          4.05E+3 (1.36E+2) $<$ \\ \cline{3-6}
& & $|f-f^*|$ &          5.64E-7 (1.47E-7) &          4.64E-7 (2.05E-7) \ph &          5.26E-7 (2.15E-7) \ph \\
& & Lower FEs &          1.88E+5 (8.38E+3) &          8.11E+5 (5.65E+4) $>$ &          2.50E+6 (8.98E+4) $>$ \\ \hline
\multirowcell{4}{SMD2} & \cbox{4}{Conflicting, convex at both levels}
& $|F-F^*|$   & \textbf{1.00E-6 (0.00E+0)} & \textbf{1.00E-6 (0.00E+0)} \ph & \textbf{1.00E-6 (0.00E+0)} \ph \\
& & Upper FEs &          1.53E+4 (2.97E+3) &          7.07E+3 (3.53E+2) $<$ &          4.41E+3 (2.50E+2) $<$ \\ \cline{3-6}
& & $|f-f^*|$ &          9.70E-5 (1.01E-4) &          5.82E-7 (2.20E-7) \ph &          4.78E-7 (2.55E-7) \ph \\
& & Lower FEs &          3.75E+5 (6.79E+4) &          6.52E+5 (3.68E+4) $>$ &          2.73E+6 (1.51E+5) $>$ \\ \hline
\multirowcell{4}{SMD3} & \cbox{4}{Cooperative, convex u.l., multimodal l.l.}
& $|F-F^*|$   & \textbf{1.00E-6 (0.00E+0)} & \textbf{1.00E-6} (3.26E-1) \ph & \textbf{1.00E-6 (0.00E+0)} \ph \\
& & Upper FEs &          8.02E+3 (1.31E+4) &          7.91E+3 (1.30E+5) \ph &          3.66E+3 (1.75E+2) $<$ \\ \cline{3-6}
& & $|f-f^*|$ &          4.14E-7 (4.16E-7) &          6.77E-7 (3.42E-1) \ph &          5.67E-7 (1.54E-7) \ph \\
& & Lower FEs &          2.65E+5 (6.80E+5) &          6.59E+5 (4.59E+6) \ph &          2.28E+6 (1.15E+5) $>$ \\ \hline
\multirowcell{4}{SMD4} & \cbox{4}{Conflicting, convex u.l., multimodal l.l.}
& $|F-F^*|$   & \textbf{1.00E-6} (1.13E+0) & 2.49E+0 (1.36E+0) \ph & 1.00E+0 (4.28E-1) \ph \\
& & Upper FEs &          3.92E+5 (4.98E+5) &          1.03E+4 (1.13E+3) \ph &          1.26E+4 (8.46E+3) \ph \\ \cline{3-6}
& & $|f-f^*|$ &          3.80E-4 (1.19E+0) &          6.95E+0 (3.96E+0) \ph &          3.65E+0 (2.89E+0) \ph \\
& & Lower FEs &          7.22E+6 (9.15E+6) &          8.22E+5 (9.46E+4) \ph &          7.71E+6 (5.22E+6) \ph \\ \hline
\multirowcell{4}{SMD5} & \cbox{4}{Conflicting, convex u.l., Rosenbrock l.l.}
& $|F-F^*|$   & \textbf{1.00E-6 (0.00E+0)} & \textbf{1.00E-6 (0.00E+0)} \ph & \textbf{1.00E-6 (0.00E+0)} \ph \\
& & Upper FEs &          1.08E+4 (1.11E+3) &          7.15E+3 (1.17E+3) $<$ &          4.87E+3 (1.72E+2) $<$ \\ \cline{3-6}
& & $|f-f^*|$ &          1.19E-6 (1.81E-7) &          1.43E-6 (9.87E-7) \ph &          9.53E-7 (4.86E-7) \ph \\
& & Lower FEs &          3.02E+5 (2.80E+4) &          5.91E+5 (9.83E+4) $>$ &          3.00E+6 (1.15E+5) $>$ \\ \hline
\multirowcell{4}{SMD6} & \cbox{4}{Unique u.l. $\ys$, infinite l.l. solution set}
& $|F-F^*|$   & 1.10E+0 (3.39E+0) & \textbf{1.00E-6 (0.00E+0)} \ph & \textbf{1.00E-6 (0.00E+0)} \ph \\
& & Upper FEs &          5.29E+5 (1.17E+4) &          4.72E+3 (3.82E+2) \ph &          3.18E+3 (1.18E+2) \ph \\ \cline{3-6}
& & $|f-f^*|$ &          3.91E+0 (1.68E+1) &          2.98E-7 (1.96E-7) \ph &          4.30E-7 (1.65E-7) \ph \\
& & Lower FEs &          9.47E+6 (1.13E+4) &          5.06E+5 (3.89E+4) \ph &          1.96E+6 (6.58E+4) \ph \\ \hline
\multirowcell{4}{SMD7} & \cbox{4}{Conflicting, multi\-modal u.l., convex l.l.}
& $|F-F^*|$   & \textbf{1.00E-6 (0.00E+0)} & 1.07E-1 (1.28E-1) \ph & 1.46E-1 (4.41E-1) \ph \\
& & Upper FEs &          7.13E+4 (1.31E+5) &          2.55E+5 (2.77E+4) \ph &          1.27E+4 (1.25E+4) \ph \\ \cline{3-6}
& & $|f-f^*|$ &          5.17E-5 (8.45E-5) &          2.44E+2 (4.26E+2) \ph &          1.31E+2 (4.09E+2) \ph \\
& & Lower FEs &          1.61E+6 (3.00E+6) &          9.74E+6 (2.21E+3) \ph &          7.68E+6 (6.31E+6) \ph \\ \hline
\multirowcell{4}{SMD8} & \cbox{4}{Conflicting, multi\-modal u.l., Rosenbrock l.l.}
& $|F-F^*|$   & \textbf{1.00E-6 (0.00E+0)} & 2.15E-5 (6.12E-6) \ph & 1.28E-2 (1.07E-2) \ph \\
& & Upper FEs &          7.35E+4 (2.28E+4) &          9.66E+4 (2.82E+2) \ph &          1.60E+4 (3.75E+1) \ph \\ \cline{3-6}
& & $|f-f^*|$ &          7.46E-5 (1.74E-4) &          3.39E-5 (1.23E-5) \ph &          1.97E-2 (1.87E-2) \ph \\
& & Lower FEs &          1.68E+6 (4.94E+5) &          9.90E+6 (7.67E+2) \ph &          9.98E+6 (0.00E+0) \ph \\ \hline
Total &  & Successes & \textbf{6 (7)} & 4 (5) \ph & 5 (5) \ph \\ \toprule
\end{tabular}
}%

\end{table*}

Table~\ref{table:smd} compares our method with BOC and BL-CMA-ES on the
$(20+20)$-dimensional SMD problems. An algorithm is considered successful if its
median upper-level accuracy, $|F-F^*|$, is less than or equal to \num{1e-6},
indicating convergence in at least 50\% of runs. As upper-level accuracies below
the termination threshold of \num{1e-6} are not meaningful, we report them as
\num{1e-6} for ease of comparison. Thus, if the median upper-level accuracy is reported as
\num{1e-6}, an interquartile range (IQR) of zero indicates convergence in at
least 75\% of runs. Note that comparing FE statistics is meaningful only when
all runs converge.  %

All methods failed on SMD4, though URA-CMA-ES failed only occasionally, whereas BOC and BL-CMA-ES failed consistently, suggesting greater robustness of URA-CMA-ES. The failure likely stems from the multimodal lower-level objective, possibly exacerbated by inter-level conflict. Such cases can often be addressed by tuning the lower-level population size, either manually or via a strategy like IPOP~\cite{auger2005}. In URA-CMA-ES, increasing $\ny$, the number of URA configurations, offers another remedy.

Apart from SMD4, our method succeeded on all problems except SMD6. This failure arises from difficulty in selecting a unique optimum from an infinite set of lower-level optimal responses. Our upper-level search models a distribution over upper-level variables only, leaving $\ysx$ fixed, whereas BOC and BL-CMA-ES learn a joint distribution and can directly select from multiple possible optimal responses $\ysx$ at the upper level. Similar limitations have been observed in some surrogate-based approaches~\cite{jiang2023}.

BL-CMA-ES also failed on SMD7 and SMD8, which have upper-level multimodality, but succeeded on SMD3, which has lower-level multimodality. BOC, by contrast, failed on all three. The absence of a collaboration mechanism in BL-CMA-ES likely makes it more robust to challenging lower-level landscapes, enabling it to solve SMD3, whereas BOC's coevolution-based collaboration appears problematic under lower-level multimodality. Both BOC and BL-CMA-ES share the same upper-level distribution mismatch issue, explaining their common failures under upper-level multimodality. URA-CMA-ES largely overcomes these issues through warm-starting-facilitated selective information sharing and a distributionally correct upper-level search, demonstrating improved versatility in handling multimodality at both levels.

Regarding computational cost, URA-CMA-ES's heuristics incur more upper-level FEs, but the upper-level overhead is negligible since the number of lower-level FEs is significantly larger. This holds when the cost of a single upper-level FE is comparable to that of a lower-level FE, as in these benchmarks. When upper-level FEs are substantially more expensive, this overhead becomes a limitation; improving the FE efficiency of our heuristics is a direction for future work. At the lower level, our method is significantly more efficient, requiring fewer FEs than both BOC (except on SMD2) and BL-CMA-ES. Finally, while BL-CMA-ES's lack of information sharing makes it the least computationally efficient method, it also grants greater flexibility than BOC.

\subsection{Results on WRA}\label{ss:results_wra}
\begin{table*}[t]
\caption{
\boldmath
Median values and IQRs (in parentheses) of final objective function
accuracy and number of function evaluations (FEs) for the $(20+20)$-dimensional WRA test suite,
aggregated over 20 runs. Significance indicators ($<$, $>$, $=$) relative to
URA-CMA-ES and success counts are determined using the same methodology
described in Table~\ref{table:smd}.
}
\label{table:wra}
\centering
{
\begin{tabular}{ccccccc}
\bottomrule
& Comment & & URA-CMA-ES (ours) & BOC \ph & BL-CMA-ES \ph \\ \hline
\multirowcell{2}{WRA1} & \cbox{2}{Bilinear} &
    $|F-F^*|$ & \textbf{1.00E-6 (0.00E+0)} & 1.25E+0 (2.10E-1) \ph & 5.28E-2 (5.51E-2) \ph \\
& & Total FEs & 5.62E+5 (9.78E+4) & 1.00E+7 (8.44E+2) \ph & 1.00E+7 (4.22E+3) \ph \\ \hline
\multirowcell{2}{WRA2} & \cbox{2}{Contains bilinear term} &
    $|F-F^*|$ & \textbf{1.00E-6 (0.00E+0)} & 6.83E-1 (9.91E-2) \ph & 6.70E-2 (5.82E-2) \ph \\
& & Total FEs & 4.81E+5 (4.04E+4) & 1.00E+7 (6.44E+2) \ph & 1.00E+7 (9.02E+3) \ph \\ \hline
\multirowcell{2}{WRA3} & \cbox{2}{Bilinear + convex term in $\xx$} &
    $|F-F^*|$ & \textbf{1.00E-6 (0.00E+0)} & 3.03E-2 (1.25E-2) \ph & 1.74E-2 (5.12E-3) \ph \\
& & Total FEs & 7.76E+5 (3.46E+4) & 3.38E+5 (1.03E+4) \ph & 1.00E+7 (3.25E+1) \ph \\ \hline
\multirowcell{2}{WRA4} & \cbox{2}{Very large optimal response set} &
    $|F-F^*|$ & 8.51E+1 (5.46E+0) & 9.45E+0 (1.23E+0) \ph & 6.74E-3 (8.04E-2) \ph \\
& & Total FEs & 1.00E+7 (2.00E+3) & 1.00E+7 (5.47E+2) \ph & 1.00E+7 (3.08E+6) \ph \\ \hline
\multirowcell{2}{WRA5} & \cbox{2}{Strictly convex-concave} &
    $|F-F^*|$ & \textbf{1.00E-6 (0.00E+0)} & \textbf{1.00E-6 (0.00E+0)} \ph & \textbf{1.00E-6 (0.00E+0)} \ph \\
& & Total FEs & 1.97E+5 (2.02E+4) & 3.43E+5 (1.44E+4) $>$ & 1.65E+6 (6.73E+4) $>$ \\ \hline
\multirowcell{2}{WRA6} & \cbox{2}{Strictly convex-concave} &
    $|F-F^*|$ & \textbf{1.00E-6 (0.00E+0)} & \textbf{1.00E-6 (0.00E+0)} \ph & \textbf{1.00E-6 (0.00E+0)} \ph \\
& & Total FEs & 4.39E+5 (9.05E+4) & 3.86E+5 (9.50E+3) $=$ & 2.58E+6 (8.37E+4) $>$ \\ \hline
\multirowcell{2}{WRA7} & \cbox{2}{Contains bilinear term} &
    $|F-F^*|$ & \textbf{1.00E-6 (0.00E+0)} & 5.10E-5 (4.39E-6) \ph & 2.64E-6 (5.72E-6) \ph \\
& & Total FEs & 4.46E+5 (9.97E+4) & 1.00E+7 (4.75E+2) \ph & 1.00E+7 (9.58E+3) \ph \\ \hline
\multirowcell{2}{WRA8} & \cbox{2}{Strictly convex-concave} &
    $|F-F^*|$ & \textbf{1.00E-6 (0.00E+0)} & \textbf{1.00E-6 (0.00E+0)} \ph & \textbf{1.00E-6 (0.00E+0)} \ph \\
& & Total FEs & 4.87E+5 (9.22E+4) & 4.11E+5 (1.41E+4) $=$ & 2.69E+6 (7.12E+4) $>$ \\ \hline
\multirowcell{2}{WRA9} & \cbox{2}{Multimodal in $\yy$} &
    $|F-F^*|$ & \textbf{1.00E-6 (0.00E+0)} & 1.80E-1 (2.29E-1) \ph & \textbf{1.00E-6 (0.00E+0)} \ph \\
& & Total FEs & 3.24E+6 (4.30E+6) & 1.00E+7 (6.21E+2) \ph & 2.80E+6 (1.12E+6) $=$ \\ \hline
\multirowcell{2}{WRA10} & \cbox{2}{Concave in both $\xx$ and $\yy$} &
    $|F-F^*|$ & 3.91E+1 (3.27E+1) & \textbf{1.00E-6 (0.00E+0)} \ph & \textbf{1.00E-6 (0.00E+0)} \ph \\
& & Total FEs & 1.00E+7 (5.19E+2) & 7.95E+5 (4.98E+4) \ph & 1.61E+6 (3.81E+4) \ph \\ \hline
\multirowcell{2}{WRA11} & \cbox{2}{Ill-conditioned} &
    $|F-F^*|$ & \textbf{1.00E-6 (0.00E+0)} & \textbf{1.00E-6 (0.00E+0)} \ph & \textbf{1.00E-6 (0.00E+0)} \ph \\
& & Total FEs & 3.33E+5 (5.19E+4) & 2.90E+6 (4.97E+5) $>$ & 8.16E+6 (2.60E+5) $>$ \\ \hline
Total &   & Successes  & \textbf{9 (9)} \ph& 5 (5) \ph & 6 (6) \ph\\ \toprule
\end{tabular}
}%
\end{table*}

Table~\ref{table:wra} compares the proposed method with BOC and BL-CMA-ES on the $(20+20)$-dimensional WRA problem sets. As these are min-max optimization problems, $f(\xx,\yy) = -F(\xx,\yy)$ in their bilevel formulation, so the upper and lower accuracies $|F-F^*|$ and $|f-f^*|$ are identical. Moreover, since the computational costs of the upper and lower objectives are identical, we report only the total number of FEs.

No method solved WRA4. In addition, BOC and BL-CMA-ES failed on WRA1, WRA2, WRA3, and WRA7; BOC also failed on WRA9, while URA-CMA-ES failed only on WRA10. We interpret these results as follows:

\emph{WRA4:} This function has a large set
($2^{d_y}\approx10^6$) of lower-level optimal responses, making it intractable
regardless of solver.

\emph{WRA1, WRA2, WRA3, WRA7:} All of these functions contain a bilinear
term $\xx\T\yy$. Combined with box constraints, this term's optimal response
$\ysx$ corresponds to one of the $2^{d_y}\approx10^6$ vertices of the box,
determined by the sign of each element of $\xx$. As randomly generated
upper-level candidate solutions differ in their signs, the marginalization
mechanism of BL-CMA-ES and BOC and the lack of information sharing across
generations cause these algorithms to initialize the lower-level solvers with an
uninformative average over these vertices at every upper-level iteration, stalling the search and preventing
convergence. Moreover, BOC's indiscriminate information sharing within lower-level generations is ineffective, since
each solver needs to select a different vertex.

\emph{WRA9:} BOC's consistent failure on this problem reflects the same
weakness on multimodality observed on the SMD benchmark.

\emph{WRA10:} The failure in convergence of URA-CMA-ES occurs because a single iteration of
improvement ($\cmax=1$) is insufficient to accurately judge convergence of the
lower-level solver~\cite{akimoto2022}. Using a larger $\cmax$ would enable
URA-CMA-ES to solve this problem at the expense of increased FEs.

On the WRA benchmark suite, URA-CMA-ES clearly demonstrates superior
versatility, particularly on functions with strong interaction. As expected, it
was also significantly more efficient than BL-CMA-ES. Though less pronounced, an
advantage in efficiency was also observed over BOC.  Of the four functions
solved by both URA-CMA-ES and BOC, BOC was faster by a slim margin on WRA6 and WRA8,
which are strictly convex-concave and well-conditioned, though its superiority
on these two functions did not meet the threshold for statistical significance.
Conversely, URA-CMA-ES was statistically significantly faster on the strongly
convex-concave WRA5, albeit only by a slight amount, and substantially faster
(roughly an order of magnitude) on the ill-conditioned WRA11. These results
suggest that the improved versatility of URA-CMA-ES contributes to a meaningful
speedup on harder problems, while on the easier, well-behaved functions (WRA5,
WRA6, WRA8) the two algorithms perform comparably.

\subsection{Ablation Study}
We perform an ablation study to elucidate the contributions of early stopping and warm starting on the performance of URA-CMA-ES. We adopt the same experimental settings described in Section~\ref{ss:exp_settings}, and report accuracy and FE statistics over 20 runs.
\subsubsection{Early Stopping}
\begin{figure}[t]
    \centering
    \includegraphics[width=0.7225\linewidth]{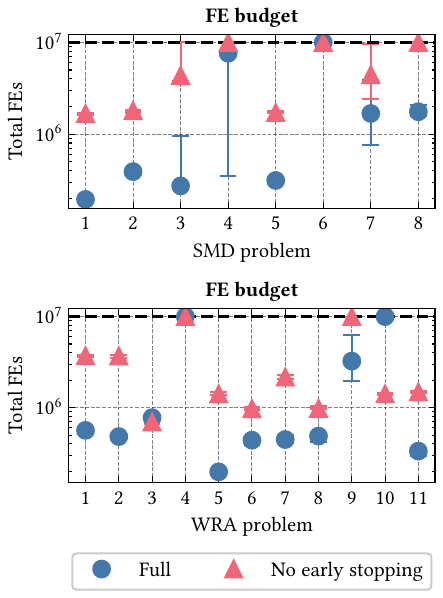}
    \caption{Total function evaluations (FEs) until convergence on SMD and WRA benchmarks for full and early-stopping-ablated URA-CMA-ES. Points and error bars indicate medians and IQRs over 20 runs, respectively.}
    \label{fig:es_ablation}
\end{figure}
To ablate the early stopping mechanism, we remove the break statement of
 Algorithm~\ref{alg:WRA} triggered when ${\tau>\tthr}$.
Figure~\ref{fig:es_ablation} shows that early stopping substantially reduces the FE cost of URA-CMA-ES on all SMD problems, and is in fact required for convergence on SMD4. A similar trend is observed on the WRA test suite, except for WRA10, where only the ablated variant converges, and WRA3, where the ablated algorithm is only very slightly faster. As discussed in Section~\ref{ss:results_wra}, failure on WRA10 with early stopping occurs when the hyperparameter $\cmax$ is set too low, resulting in inaccurate convergence detection. These results suggest that early stopping is highly effective for accelerating search in most cases, but can be disabled or tuned via $\cmax$ if convergence issues arise.

\subsubsection{Warm Starting}\label{ss:ablate_ws}
\begin{figure}[t]
    \centering
    \includegraphics[width=0.7225\linewidth]{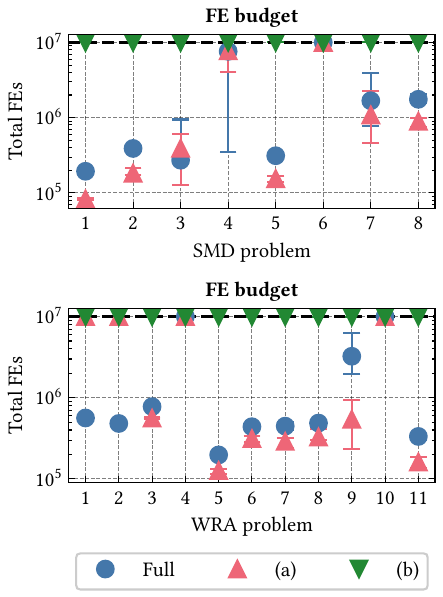}
    \caption{Total function evaluations (FEs) until convergence on SMD and WRA benchmarks for full URA-CMA-ES and the two types of warm-starting ablation.
    Points and error bars indicate medians and IQRs over 20 runs, respectively.}
    \label{fig:ws_ablation}
\end{figure}
We consider two ways of ablating the warm starting
mechanism: (a) setting the number of configurations to $\ny=1$, and (b) setting
$\ny=1$ while also refreshing the lower-level configuration after each
upper-level iteration. Ablation (a) forces all lower-level solvers in the subsequent
upper-level generation to share the same initial configuration, namely the
configuration of the best-performing lower-level solver from the previous
generation. This is qualitatively similar, though not strictly identical, to the
restriction that all lower-level solvers of BL-CMA-ES and BOC are initialized by
marginalizing the upper-level distribution. Ablation (b) is a much stronger
restriction that completely eliminates all information sharing across generations.

Figure~\ref{fig:ws_ablation} shows the impact of warm-starting ablation on the performance of the proposed method. Ablation~(a) demonstrates that warm starting never impedes URA convergence and incurs only slight additional overhead in most cases. Furthermore, warm starting proves necessary for WRA1 and WRA2, where maintaining multiple configurations facilitates efficient traversal of box-constraint vertices to locate the lower-level optimal response when bilinear terms dominate. Given these benefits and the absence of significant drawbacks, we recommend incorporating warm starting in all cases. Finally, ablation~(b) fails to solve any benchmark problem in either test suite, highlighting that preserving some information from the previous generation, even if limited and nonspecific to each upper-level candidate, is essential for convergence.

\section{Conclusion}
We introduced URA-CMA-ES, an upper-level value-function-based approach incorporating warm
starting and early stopping to address key limitations in prior black-box
bilevel methods, and demonstrated its competitiveness on two challenging
benchmark suites. Its selective information sharing provides
acceleration over BL-CMA-ES comparable to that of BOC without sacrificing generality. Moreover, URA-CMA-ES overcomes other lower-level limitations and offers greater versatility than both methods, particularly in settings with multimodality and strong inter-variable interaction.

Several directions exist for future work to extend our method. The URA algorithm generalizes naturally
to other rank-based evolutionary algorithms, enabling optimization of discrete
or mixed-variable BLOPs. Since our method does not require a specific
lower-level solver, integration with specialized local search or white-box
algorithms where applicable would further broaden its
applicability. Finally, further investigation of the early stopping and
warm-starting heuristics could improve budget efficiency.

\begin{acks}
The authors would like to express their sincere gratitude to the authors of \cite{huang2023} for generously providing the source code of the BOC algorithm used in this study.
\end{acks}

\bibliographystyle{ACM-Reference-Format}
\bibliography{gecco2026}
\end{document}